\documentclass[11pt]{article}

\usepackage{graphicx}
\usepackage{booktabs}
\usepackage{amsmath,amssymb}
\usepackage{natbib}
\usepackage{hyperref}
\usepackage{xcolor}
\usepackage[
  top=0.8in,
  bottom=0.8in,
  left=0.85in,
  right=0.85in
]{geometry}

\title{Beyond Accuracy: A Multidimensional Evaluation of Statistical Reasoning in Large Language Models}

\author{
Monnie McGee\\
Department of Statistics and Data Science\\
Southern Methodist University
\and
Mateo Langston Smith\\
Research Technology Services\\
Southern Methodist University
\and
Julian Cabrera\\
Department of Statistics and Data Science\\
Southern Methodist University
}

\date{}        

\begin{document}

\maketitle

\begin{center}
\small
This manuscript has been submitted for peer review. This version may differ from the final published article.
Comments are welcome.
\end{center}

\begin{abstract}
Statistical reasoning is multidimensional, yet evaluations of large language models (LLMs) typically emphasize response accuracy while overlooking how models construct and communicate statistical explanations. This study demonstrates the value of a multidimensional evaluation by combining response accuracy, response behavior, structural topic modeling, and lexical similarity analysis. The framework is applied to explanations generated by 15 current-generation LLMs responding to 90 questions drawn from four statistics examinations spanning high school, undergraduate, and graduate levels. Accuracy varied substantially across models, ranging from 55\% to 78\%. In contrast, structural topic modeling revealed a common conceptual organization of statistical reasoning across all models, while lexical similarity analysis identified modest but consistent vendor-specific differences in explanatory style. Models developed by the same vendor (e.g. Anthropic, OpenAI) produced explanations that were slightly more similar than models from different vendors. These findings demonstrate that statistical reasoning in contemporary LLMs cannot be characterized by accuracy alone and illustrate how complementary analyses of response behavior and model-generated explanations provide a more comprehensive evaluation of statistical reasoning in generative AI.
\end{abstract}


\section{Introduction}\label{sec:intro}

Large language models (LLMs) are increasingly used to answer statistical questions, explain statistical concepts, and provide individualized feedback to students \cite{KhanAcademy2024}. As their use in statistics education expands, an important question arises: how should statistical reasoning in these systems be evaluated? Existing studies have demonstrated substantial variation in performance across LLM families and model generations, motivating continued evaluation as new models are released \cite{mcgeesadler2025}. However, most evaluations focus almost exclusively on response accuracy, even though statistical reasoning also involves recognizing uncertainty, selecting appropriate concepts, and communicating explanations.

To address this limitation, we evaluate the statistical reasoning reflected in output generated by LLMs along four complementary dimensions: response accuracy, response behavior, the conceptual organization of statistical reasoning using structural topic modeling, and the lexical similarity of explanations using TF--IDF cosine similarity. Together, these measures provide a broader characterization of statistical reasoning than accuracy alone. We evaluate 15 contemporary LLMs using four assessments. Three are well-established: the 2011 AP Statistics Examination, the ACTM examination \cite{ACTMExam}, the Comprehensive Assessment of Outcomes in Statistics (CAOS) \cite{caos}. The fourth is a graduate-level examination developed by one of the authors.

This study makes three contributions. First, it demonstrates the value of evaluating statistical reasoning using multiple complementary measures rather than response accuracy alone. Second, it shows how structural topic modeling and TF–IDF cosine similarity analysis provide complementary perspectives on model-generated statistical explanations by characterizing the conceptual organization of statistical reasoning and its linguistic realization. Third, it provides a comprehensive evaluation of 15 contemporary LLMs across statistics assessments spanning high school, undergraduate, and graduate levels. Together, these complementary analyses reveal aspects of statistical reasoning that are not apparent from response accuracy alone.


\section{Background and Related Work}\label{sec:background}

The integration of large language models (LLMs) into statistics and data science education has led to a growing literature evaluating their performance on course-related tasks. Prior studies show that LLMs can correctly answer many standard statistics questions, often performing comparably to average students, particularly on well-structured or lower-complexity problems \cite{mcgeesadler2025, ellis23, williams2024}. However, performance varies with model version, prompt design  \cite{yao2024}, and task complexity \cite{ruta2025}, and even advanced models can produce errors in nuanced statistical reasoning, code generation, and interpretation \cite{herklotz2025trustllmstutorstudents}. Although models often perform well on structured or computational tasks, they remain susceptible to errors in statistical interpretation, model selection, and conceptual reasoning.  Consequently, LLMs are generally viewed as a supplement rather than a replacement for expert instruction \cite{mcgeesadler2025, ellis23}.

A second line of research has examined LLMs as providers of formative feedback rather than problem solvers. These studies suggest that model-generated feedback can approximate expert evaluation but may contain conceptual errors and inconsistencies, particularly when explanations require deeper statistical understanding \cite{herklotz2025trustllmstutorstudents, schwarz2025}. Unlike these studies, the present work focuses on the statistical reasoning exhibited while solving problems rather than the quality of feedback generated after a solution is known.

At the same time, the generative AI landscape has evolved rapidly, with new model generations introduced across multiple systems. Consequently, many published evaluations examine models that have since been superseded. Moreover, much of the comparative literature has been conducted outside statistics education, particularly in biomedical applications \cite{Nascimento2024LLM4DSEL}. As a result, there is limited empirical evidence regarding the performance of current-generation LLMs on standardized statistics assessments. 

Although recent studies have begun to examine the text of LLM-generated explanations, such analyses remain comparatively rare relative to accuracy-based evaluations. McGee and Sadler \cite{mcgeesadler2025} extended this literature by analyzing the text of model-generated explanations using topic modeling in addition to response accuracy, illustrating the potential value of examining explanation content alongside correctness. The present study extends that work in three important ways. First, it evaluates a substantially broader set of contemporary LLMs across multiple platforms. Second, it integrates response behavior, structural topic modeling, and lexical similarity into a unified evaluation of model-generated explanations. Third, it compares models spanning free and paid access tiers using common standardized statistics assessments. Together, these complementary analyses provide a more comprehensive characterization of statistical reasoning than response accuracy alone and, in doing so, yield an updated comparison of contemporary LLMs.

\section{Methods}\label{sec:methods}


\subsection{Data Collection}
\label{sec:data_collection}

Responses were collected from 15 large language models representing four
model families: three Google Gemini models (2.0, 2.5, and 3.1), three xAI
Grok models (2-vision, 4.1, and 4.3), six OpenAI models (\mbox{GPT-4o},
GPT-5, GPT-5.1, GPT-5.4-nano, GPT-5.4-mini, and GPT-5.4), and three
Anthropic Claude models (3.7, 4, and 4.6). All models were queried via
their respective APIs in a zero-shot format: each question was presented
individually without examples, prior context, or conversational history.
Each model received the complete question, including answer choices for multiple-choice items and any accompanying images. Each response consisted of both an answer and an explanation. 

Data collection occurred in two phases. Fourteen models were queried before May 15, 2026. Grok 4.3 was evaluated after its public release and replaced the retired Grok 2 and Grok 4.1 models, whose archived responses were retained for comparison; therefore, results for Grok~2 and Grok~4.1 reflect responses generated prior to their retirement and cannot be reproduced via the API. No conversational memory was retained between questions, and each response was generated independently.

\subsection{Instruments}
\label{sec:instruments}

Responses were collected for questions drawn from four statistics assessments
spanning a range of educational levels. The \textit{Comprehensive Assessment
of Outcomes in Statistics} \cite[CAOS]{caos} is a 40-item multiple-choice (MC)
instrument designed to assess conceptual understanding of introductory
statistics after a first college course. The \textit{American College Testing
Mathematics State Statistics Exam} \cite[ACTM]{ACTMExam} is a 28-item
instrument consisting of 25 MC and 3 short answer (SA) questions, designed for high school
students. The \textit{Advanced Placement Statistics Exam} \cite[AP]{cb2011}
contributes 6 SA questions drawn from a college-credit-bearing exam at the
introductory statistics level. The \textit{Graduate Exam} (GE) is a 16-item
instrument, with 8 MC and 8 SA questions, developed for first-year doctoral students. These four exams were
selected to span educational levels from high school through doctoral training; they are not the specific object of inference and are treated as a
convenience sample of statistics assessments at different educational levels.

Across all four exams there are 90 questions total: 73 MC and 17 SA.
Table~\ref{tab:exam_structure} summarizes the structure of each exam. Of the
73 MC questions, 24 involve images that must be interpreted to answer
correctly: 20 in CAOS, 2 in ACTM, and 2 in the GE. An additional
variable, \texttt{cross\_reference}, was coded to indicate whether a
question explicitly references context, data, or results from another
question. Cross-referential questions were identified in CAOS (23 questions)
and ACTM (7 questions). In CAOS, cross-referential questions belong to item
sets in which multiple questions share a common scenario or image, with all
necessary information present in the shared stem. In ACTM, cross-referential
questions require results computed in a prior question that are not
restated, making them unanswerable in the zero-shot single-question format
used here. This distinction and its consequences for model behavior are
discussed further in Section~\ref{sec:results}.

\begin{table}[ht]
\caption{Structure of the four statistics assessments used in the study.}
\label{tab:exam_structure}
\begin{tabular}{lrrrrr}
\hline
Exam   & Level          & Total & MC & SA & Image (MC) \\
\hline
CAOS   & Intro college  & 40    & 40 &  0 & 20         \\
ACTM   & High school    & 28    & 25 &  3 &  2         \\
GE     & Doctoral       & 16    &  8 &  8 &  2         \\
AP     & Intro college  &  6    &  0 &  6 &  0         \\
\hline
Total  &                & 90    & 73 & 17 & 24         \\
\hline
\end{tabular}
\end{table}

\subsection{Answer Key Verification}
\label{sec:key_verification}

Prior to analysis, all MC answer keys were verified against model consensus
responses. Questions on which zero or one model answered correctly were
flagged for manual review against the original exam materials. This process
identified two confirmed key errors in the ACTM exam: question~6, for which
the keyed answer (d: \textit{neither type of inference}) was incorrect and
was corrected to \textit{a: population inference}; and question~13, for
which the keyed answer (a: \textit{simple random, stratified, convenience})
was incorrect and was corrected to \textit{c: stratified, simple random,
convenience}. No key errors were identified in the CAOS or GE exams.
Remaining near-zero accuracy items reflected genuinely difficult questions, image-dependent items, or ACTM cross-referential questions that could not be answered in the zero-shot setting.

\subsection{Accuracy Coding}
\label{sec:accuracy_coding}

For MC questions, model responses were coded as correct (1) or incorrect (0)
by comparing each response to the verified answer key. Two response types
required special treatment. Responses coded as \textit{NC} (no conclusion)
indicate that a model recognized the question required information it could
not access, most commonly an image it could not interpret or data from a
prior question. These responses indicate that the model explicitly recognized insufficient information and therefore declined to provide a definitive answer.
Responses coded as \textit{NA} indicate that no response was produced,
either due to an API error or a content filter. Both NC and NA responses
were treated as incorrect (0) for accuracy analysis, as neither constitutes
a correct answer regardless of the reason. The prevalence of NC and NA
responses by model and exam is reported in Section~\ref{sec:results} as an
analytically meaningful finding in its own right.

Accuracy scores are available for all 73 MC questions across 15 models,
yielding 1,095 documents with known accuracy. Accuracy for the 17 SA
questions requires human scoring using a standardized rubric and is deferred
to a follow-up study.

\subsection{Structural Topic Modeling}
\label{sec:stm}

The explanatory text accompanying each model response was analyzed using
Structural Topic Modeling (STM) \cite{roberts2019}. STM is a
generalization of Latent Dirichlet Allocation (LDA)
\cite{blei2003} that incorporates document-level covariates, allowing
topic prevalence to vary systematically with observed document
characteristics.

Each explanation was treated as a separate document, yielding a corpus of
1,350 documents (90 questions answered by each of 15 LLMs). Document-level
metadata included the name of the LLM, the name of the examination, question type (multiple choice or
short answer), presence of an image (Yes or No), question identifier, and, for
multiple-choice responses, the response outcome (correct, incorrect, no
conclusion, or no response).

A STM was fit to the complete corpus rather than fitting separate models for each LLM. Pooling produced a common topic space that permitted direct comparison of topic prevalence across models while avoiding unstable estimation from the approximately 90 documents available for each individual LLM. Two STM analyses were conducted. The primary model included all 1,350
documents, with topic prevalence modeled as a function of LLM,
examination, question type, and image presence. A secondary analysis was
restricted to the 1,095 multiple-choice responses and included response
outcome (correct, incorrect, no conclusion, or no response) as an
additional prevalence covariate to investigate whether different response
outcomes were associated with distinct patterns of statistical reasoning. Results for the single pooled STM are reported because there was no evidence of differences in topic prevalence when the response outcome was consider.  

Following model estimation, each topic was assigned an interpretable label through qualitative review of the STM output. Topic interpretation was based on three complementary sources of information: (1) the highest-probability words associated with each topic, (2) FREX words, which balance word frequency and exclusivity to identify terms that are both common within a topic and distinctive from other topics, and (3) representative documents having high posterior probability for each topic. Topic labels were assigned iteratively by examining these sources together, with emphasis placed on the representative documents to ensure that labels reflected the underlying statistical reasoning rather than isolated keywords or question-specific terminology. Labels therefore describe common reasoning patterns (e.g., Confidence Interval Interpretation, Sampling Bias and Generalizability, and Procedural Statistical Computation) rather than merely the subject matter of individual examination questions.

The number of topics ($K$) was selected using the
\texttt{searchK()} function in the \texttt{stm} package
\cite{roberts2019}. Candidate values from $K=12$ to $K=22$
were evaluated using held-out likelihood, semantic coherence,
exclusivity, and residual dispersion. The final value of $K = 20$
was chosen by balancing these quantitative diagnostics with the
interpretability of the resulting topics.

\subsubsection{Topic Prevalence Model}

For the primary STM, topic prevalence was modeled as a function of LLM,
examination, question type, and image presence. The examination variable
was included to account for differences in content and educational level
across assessments, whereas the primary inferential interest was in
differences among LLMs.

The secondary STM, restricted to multiple-choice responses, additionally
included response outcome (correct, incorrect, no conclusion, or no
response) as a prevalence covariate. Topic effects were estimated using the
\texttt{estimateEffect()} function in the \texttt{stm} package, which accounts
for uncertainty in the estimated topic proportions when evaluating
covariate effects.

\subsection{Lexical Similarity Analysis}

Structural topic modeling characterizes the statistical concepts discussed in model explanations but does not measure how similarly those concepts are expressed. To complement the STM analysis, lexical similarity was quantified using TF–IDF cosine similarity \cite{SALTON1988513, manning2008introduction}, providing a measure of similarity in explanatory language.

Each explanation was represented as a term frequency–inverse document frequency (TF–IDF) vector constructed from the cleaned corpus after tokenization, stop-word removal, and term-frequency trimming. Cosine similarity was then computed between every pair of document vectors, producing a similarity score ranging from 0 (no shared vocabulary) to 1 (identical TF–IDF representations). For each examination question, pairwise similarities were calculated for all model pairs, yielding $\binom{15}{2} = 105$ pairwise comparisons per question and $9,450$ comparisons across the 90 questions.

Average pairwise similarities were summarized to produce a model-by-model similarity matrix. Relationships among models were further explored using metric multidimensional scaling (MDS) based on the cosine distance ($1-$cosine similarity), allowing models with similar explanatory language to be displayed in a two-dimensional space. To investigate whether models developed by the same vendor exhibited more similar explanatory styles than models from different vendors, pairwise comparisons were classified as either within-vendor or between-vendor. Mean similarities for these two groups were estimated using a nonparametric bootstrap in which examination questions were resampled with replacement, preserving the dependence among model pairs within each question. The bootstrap procedure was used to estimate the mean within-vendor and between-vendor similarities as well as a 95\% confidence interval for their difference.

\subsection{Software \& Generative AI Use}
\label{sec:software}

All analyses were conducted in \texttt{R} \cite[version 4.6.0]{rcore}
using the \texttt{stm} \cite{roberts2019}, \texttt{quanteda} \cite{quanteda},
and \texttt{readtext} \cite{readtext} packages. Generative artificial intelligence was used as an editorial and programming assistant during the preparation of this manuscript. ChatGPT (OpenAI GPT-5.5) was used to assist with drafting and revising text, developing and debugging R code, exploring alternative approaches to Structural Topic Modeling, and improving the clarity and organization of the manuscript. All statistical methods, analytical decisions, implementation of the analyses, interpretation of results, and scientific conclusions were determined independently by the authors. All code was executed and verified by the authors, and every result reported in the manuscript was independently validated. The authors assume full responsibility for the content of this article.

\section{Results}\label{sec:results}

We first summarize model performance using conventional measures of response accuracy and response behavior. We then examine the statistical reasoning reflected in model-generated explanations using structural topic modeling and lexical similarity analysis.

\subsection{Accuracy}\label{sec:accuracy}

We begin by comparing model accuracy on the three multiple-choice examinations. The AP Statistics examination was excluded from this analysis because it consists entirely of free-response items. Table \ref{tab:mc_accuracy_model_exam} summarizes the percentage of correct responses for each model on the ACTM, CAOS, and graduate-level examinations, together with an overall weighted accuracy.
\begin{table}[htbp]
\centering
\caption{Multiple-choice accuracy by model and exam. Values are percentages. Overall accuracy is computed as a weighted average using 25 ACTM questions, 40 CAOS questions, and 8 graduate exam questions.}
\label{tab:mc_accuracy_model_exam}
\begin{tabular}{lrrrr}
\toprule
Model & ACTM & CAOS & GE & Overall \\
            & $(n=25)$ & $(n=40)$ & $(n=8)$ & $(n=73)$ \\
\midrule
\multicolumn{5}{l}{\textbf{Anthropic}} \\
Claude 3.7 & 68.0 & 65.0 & 50.0 & 64.4 \\
Claude 4   & 68.0 & 67.5 & 62.5 & 67.1 \\
Claude 4.6 & 72.0 & 67.5 & 62.5 & 68.5 \\
\addlinespace

\multicolumn{5}{l}{\textbf{Google}} \\
Gemini 2.0 & 68.0 & 47.5 & 50.0 & 54.8 \\
Gemini 2.5 & 60.0 & 72.5 & 62.5 & 67.1 \\
Gemini 3.1 & 72.0 & 82.5 & 62.5 & 76.7 \\
\addlinespace

\multicolumn{5}{l}{\textbf{OpenAI}} \\
 GPT-4o        & 68.0 & 70.0 & 75.0 & 69.9 \\
GPT-5         & 64.0 & 72.5 & 50.0 & 67.1 \\
GPT-5.1       & 52.0 & 75.0 & 62.5 & 65.8 \\
GPT-5.4       & 68.0 & 80.0 & 62.5 & 74.0 \\
GPT-5.4-mini  & 68.0 & 72.5 & 62.5 & 69.9 \\
GPT-5.4-nano  & 72.0 & 65.0 & 62.5 & 67.1 \\
\addlinespace

\multicolumn{5}{l}{\textbf{xAI}} \\
Grok 2-vision & 60.0 & 62.5 & 75.0 & 63.0 \\
Grok 4.1      & 72.0 & 72.5 & 62.5 & 71.2 \\
Grok 4.3      & 84.0 & 77.5 & 62.5 & 78.1 \\
\bottomrule
\end{tabular}
\end{table}

Performance varied substantially across both models and examinations. Overall accuracy was computed by weighting the accuracy from each exam by the number of questions on the exam. Weighted accuracy ranged from 54.8\% for Gemini 2.0 to 78.1\% for Grok 4.3, with newer model generations generally outperforming earlier versions within the same model family. Accuracy was most variable on the ACTM examination, where scores ranged from 52.0\% to 84.0\%. In contrast, performance on the CAOS examination was generally higher and somewhat less variable across models, although notable differences remained. Performance on the graduate-level examination showed comparatively little variation, but interpretation is limited by the small number of multiple-choice questions (n=8).

No single model achieved the highest accuracy on every examination. Grok 4.3 obtained the highest score on the ACTM examination (84.0\%), whereas Gemini 3.1 achieved the highest score on the CAOS examination (82.5\%). These results indicate that relative model performance depends on the assessment instrument as well as the underlying model, suggesting that different examinations emphasize different aspects of statistical reasoning.

Although Table \ref{tab:mc_accuracy_model_exam} summarizes how often each model answered correctly, accuracy alone does not distinguish between incorrect answers and responses in which a model explicitly recognized insufficient information. We examine these response behaviors next.

\subsection{Response behavior}\label{sec:behavior}

To better understand incorrect responses, we classified each multiple-choice response as correct, incorrect, no conclusion, or no response. A response of ``no conclusion''' indicates that the model reasoned through the question but was unable to come to a conclusion. ``No response'' indicates that the question was unanswered (left blank). Table \ref{tab:response_behavior} summarizes the distribution of multiple-choice response outcomes across the 15 models. In addition to differences in overall accuracy, the models exhibited distinct response behaviors when they failed to produce a correct answer. The highest-performing models (Grok 4.3, Gemini 3.1, GPT-5.4, and Grok 4.1) combined high accuracy with very few responses classified as either ``no conclusion'' or ``no response.'' Grok 4.3, for example, produced no instances of either outcome.

\begin{table}[!h]
\centering
\caption{\label{tab:response_behavior}Percentage of items correct, incorrect, no conclusion (NC), and no response (NR) by model. NC indicates that the model gave some text output but did not give a definitive answer. NR indicates that the model left the response blank.}
\centering
\begin{tabular}[t]{lrrrr}
\toprule
Model & Correct & Incorrect  & NC & NR\\
\midrule
Grok43 & 78.1 & 21.9 & 0.0 & 0.0\\
Gemini31 & 76.7 & 17.8 & 5.5 & 0.0\\
GPT54 & 74.0 & 24.7 & 1.4 & 0.0\\
Grok41 & 71.2 & 26.0 & 2.7 & 0.0\\
GPT4o & 69.9 & 20.5 & 9.6 & 0.0\\
\addlinespace
GPT54-mini & 69.9 & 26.0 & 4.1 & 0.0\\
Claude46 & 68.5 & 24.7 & 6.8 & 0.0\\
Claude4 & 67.1 & 21.9 & 11.0 & 0.0\\
GPT5 & 67.1 & 9.6 & 19.2 & 4.1\\
GPT54-nano & 67.1 & 26.0 & 6.8 & 0.0\\
\addlinespace
Gemini25 & 67.1 & 26.0 & 6.8 & 0.0\\
GPT51 & 65.8 & 24.7 & 9.6 & 0.0\\
Claude37 & 64.4 & 19.2 & 16.4 & 0.0\\
Grok2-vision & 63.0 & 31.5 & 5.5 & 0.0\\
Gemini20 & 54.8 & 20.5 & 4.1 & 20.5\\
\bottomrule
\end{tabular}
\end{table}

The lower-performing models differed in the ways they failed. Some models, such as Grok 2-vision, primarily produced incorrect answers (31.5\%) while seldom declining to answer. In contrast, GPT-5 and Claude 3.7 more frequently produced responses in which the model explicitly reasoned through the problem but concluded that a definitive answer could not be determined, accounting for 19.2\% and 16.4\% of responses, respectively. Gemini 2.0 exhibited a different pattern, producing the highest rate of missing responses (20.5\%), indicating that a substantial proportion of questions yielded no usable response rather than an explicit statistical answer. These findings demonstrate that models differ not only in overall accuracy but also in how they respond when a correct answer is not produced. This distinction motivates the subsequent analyses of model-generated explanations.

\subsection{Common Conceptual Organization of Statistical Reasoning}\label{sec:stmresults}

Structural topic modeling revealed that the models shared a common conceptual organization of statistical reasoning despite substantial differences in response accuracy. Twenty topics were identified that collectively describe the conceptual organization of statistical reasoning exhibited by the models. These topics span core statistical concepts including inference, descriptive statistics, probability, regression, sampling, and graphical interpretation. The 20 topics represent the broad conceptual organization of statistical reasoning rather than isolated examination-specific vocabulary. Several topics correspond to fundamental statistical concepts encountered throughout introductory statistics, including hypothesis testing, confidence intervals, sampling distributions, randomization, and graphical interpretation.

Table \ref{tab:topic_validation} provides evidence that the identified topics represent general reasoning patterns rather than concepts unique to individual questions. Eighteen of the 20 topics were associated with multiple examination questions, and eight topics appeared across three or four examinations. For example, Topic 12 (Distribution Summaries and Boxplot Interpretation) and Topic 19 (Randomization and Causal Inference) were represented across all four examinations, whereas only Topics 4 and 20 were primarily associated with a single examination question.

\begin{table}[!ht]
\caption{Validation of STM topic labels using examination questions. For each topic, the table reports the number of exam questions with an average topic probability of at least 0.25, together with the number of examinations represented. Topics represented across multiple questions and examinations provide evidence that the STM identified general reasoning patterns rather than vocabulary unique to individual questions.}
\label{tab:topic_validation}
\centering
\small
\setlength{\tabcolsep}{3pt}
\begin{tabular}{clcc}
\toprule
\textbf{Topic} &
\textbf{Topic Label} &
\textbf{Questions} &
\textbf{Exams} \\
\midrule
1  & Robust Estimation               & 2  & 1 \\
2  & Sampling Bias and Generalizability                   & 2  & 2 \\
3  & Diagnostic Testing          & 4  & 2 \\
4  & Regression Design \& Experimental Planning       & 1  & 1 \\
5  & Regression Model Interpretation                      & 7  & 2 \\
6  & Randomized Comparisons            & 2  & 1 \\
7  & Sampling Distributions            & 8  & 2 \\
8  & Reasoning with Incomplete Information                & 8  & 2 \\
9  & Procedural Statistical Computation                   & 6  & 3 \\
10 & Probability Modeling and Simulation                  & 2  & 2 \\
11 & Sample Size, Precision, and Margin of Error          & 2  & 2 \\
12 & Boxplot Interpretation    & 12 & 4 \\
13 & Correlation and Scatterplot Interpretation           & 4  & 2 \\
14 & Histogram Interpretation      & 11 & 1 \\
15 & Hypothesis Testing    & 12 & 2 \\
16 & Association in Contingency Tables                    & 5  & 2 \\
17 & Confidence Interval Interpretation                   & 7  & 3 \\
18 & Sample Size and Precision                 & 4  & 3 \\
19 & Randomization and Causal Inference                   & 8  & 4 \\
20 & Discrete Probability Distributions                   & 1  & 1 \\
\bottomrule
\end{tabular}
\end{table}

Taken together, these results indicate that the models relied on a common conceptual organization of statistical reasoning despite substantial differences in response accuracy. The primary differences among models therefore appear to lie not in the statistical concepts invoked during explanation, but in how reliably those concepts are applied and expressed.

\subsection{Topic Prevalence Across Models}\label{sec:lexsim}

\begin{figure*}[t]
\centering
\includegraphics[width=\textwidth]{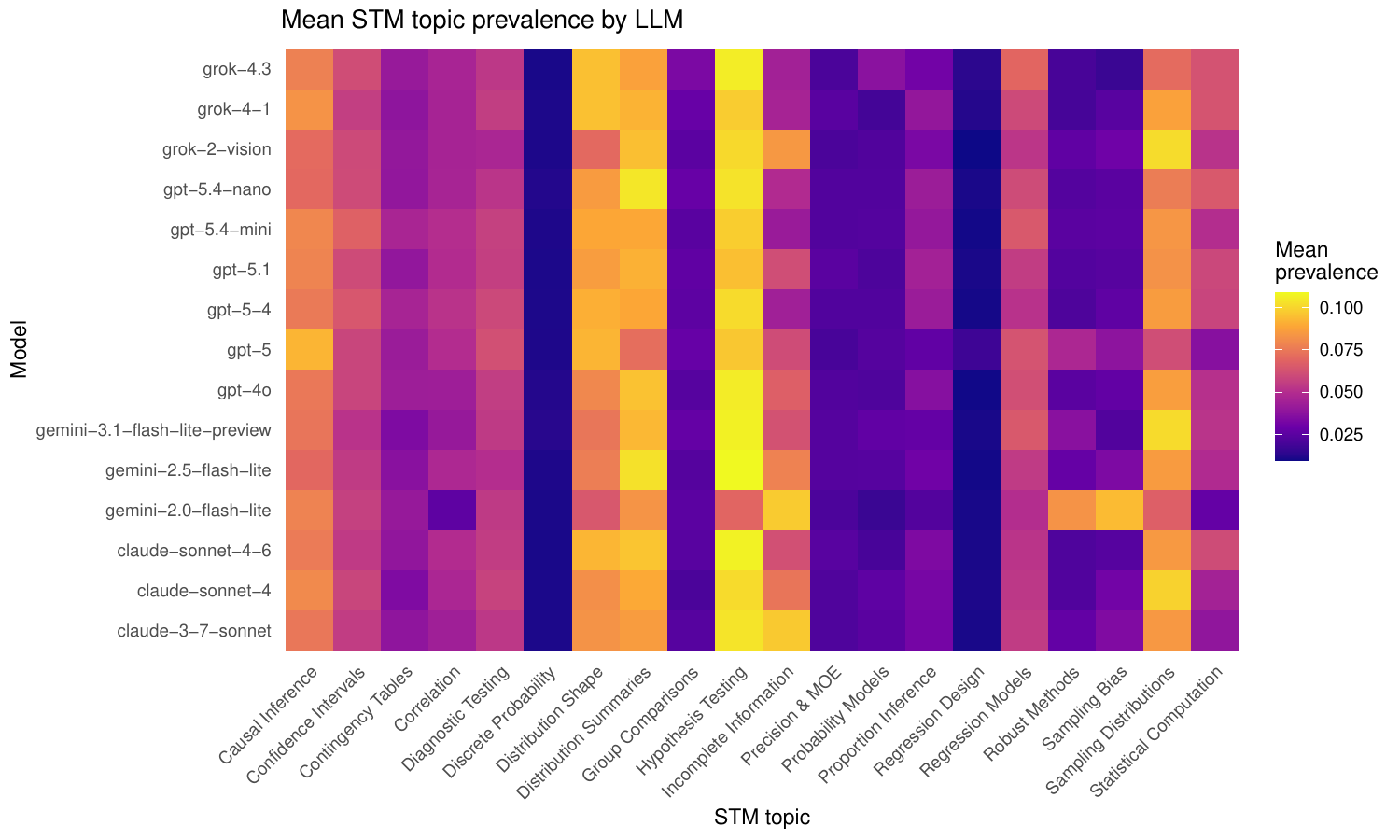}
\caption{Heatmap of mean topic prevalence by model. Rows correspond to the 15 large language models and columns correspond to the 20 topics identified by the structural topic model. Cell color represents the average posterior topic probability for a topic across all responses generated by a model. Similar color patterns across rows indicate that the models exhibit a common conceptual organization of statistical reasoning despite differences in response accuracy.}
\label{fig:topic_prevalence}
\end{figure*}

Figure \ref{fig:topic_prevalence} summarizes the average prevalence of each statistical reasoning topic for every model. The y-axis lists the 15 large language models and the 20 statistical reasoning topics identified by the STM on the x-axis. The mean topic proportion (average posterior probability) is calculated across all responses generated by a given model. Dark purple indicates a lower average prevalence, whereas yellow indicates a higher average prevalence. Since the topic proportions for each document sum to one, the color in each cell represents the average fraction of an explanation devoted to that reasoning topic.

The most striking feature of the figure is the similarity of the rows. Nearly all models exhibit the same broad pattern of topic prevalence. Topics such as Hypothesis Testing and Statistical Decision Making, Distribution Summaries and Boxplot Interpretation, Distribution Shape and Histogram Interpretation, Sampling Distributions and Standardization, and Randomization and Causal Inference are among the most prevalent across nearly every model. These results reinforce the conclusion that the models share a common conceptual organization despite differing substantially in response accuracy.

The STM results indicate that differences in model accuracy are not explained by fundamentally different conceptual organizations of statistical reasoning. We therefore examined whether the models differed instead in the language used to express those common concepts.

\subsection{Lexical Similarity}\label{sec:lexsim}

\begin{figure}[t]
\centering
\includegraphics[width=\columnwidth]{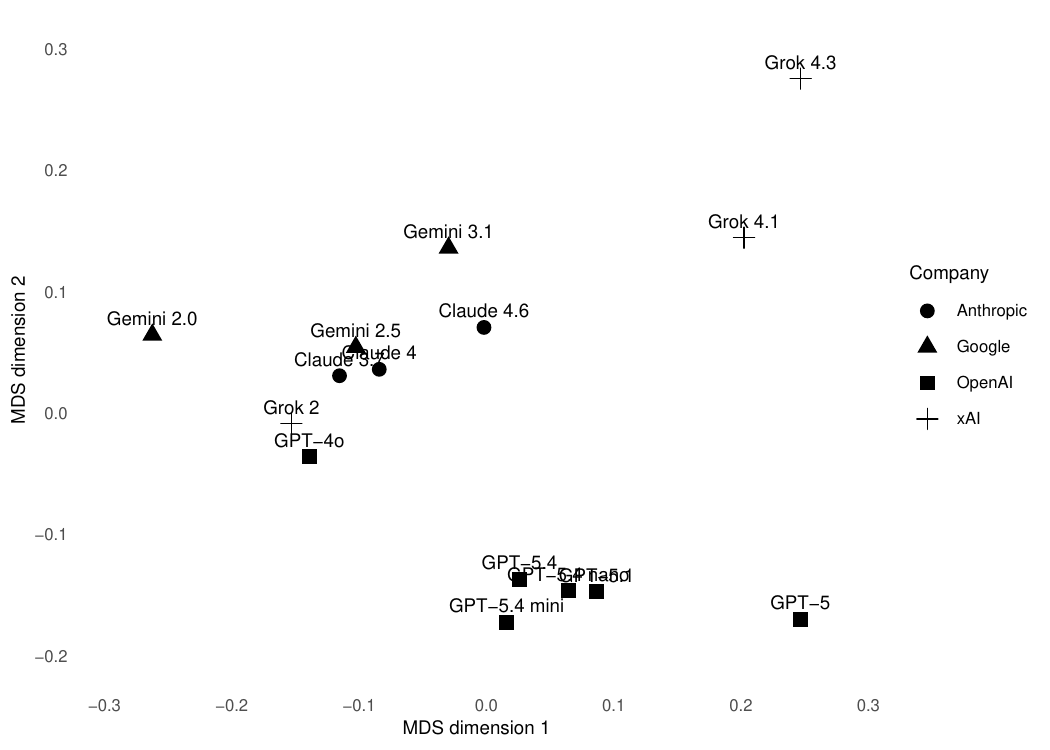}
\caption{Multidimensional scaling (MDS) representation of pairwise TF--IDF cosine dissimilarities among the 15 LLMs. Models that are closer together produced more similar explanatory language across the examination questions. Only the relative distances between points are meaningful; the orientation of the axes is arbitrary.}
\label{fig:mds}
\end{figure}

Although the STM revealed a common conceptual organization of statistical reasoning, it does not address whether models express those ideas using similar language. We therefore examined lexical similarity using TF–IDF cosine similarity. Figure~\ref{fig:mds} presents a two-dimensional multidimensional scaling (MDS) representation of the pairwise TF--IDF cosine similarity matrix. Each point represents one LLM, and the distance between two points approximates the dissimilarity of their explanations across all examination questions. Models located close together produced more similar explanatory language, whereas models farther apart exhibited greater lexical differences. The axes themselves have no substantive interpretation; only the relative distances among points are meaningful.

The MDS configuration reveals modest vendor-specific clustering rather than complete separation by company. The three Anthropic models form a compact cluster near the center of the figure, indicating highly similar explanatory language across model generations. The OpenAI models also cluster together, although GPT-5 is noticeably separated from the other OpenAI models, suggesting a distinct explanatory style. The Google models occupy an intermediate region, with Gemini 3.1 separated somewhat from Gemini 2.0 and 2.5, while the xAI models are more dispersed, reflecting greater variation across Grok generations. Despite these vendor-specific patterns, all models occupy a relatively compact region of the configuration, consistent with the moderate cosine similarities observed across all pairwise comparisons. The MDS analysis indicates that contemporary LLMs organize statistical explanations around a common set of concepts while exhibiting only modest differences in the language used to express those concepts.

Table \ref{tab:vendor_pairs} summarizes average TF--IDF cosine similarity between explanations produced by models from the same developer and from different developers. Overall, within-vendor model pairs exhibited slightly higher similarity than between-vendor pairs (0.509 versus 0.481), indicating that models from the same organization tended to produce somewhat more similar explanations. However, the difference was modest, suggesting substantial overlap in explanatory language across vendors.

\begin{table}[!h]
\centering
\caption{\label{tab:vendor_pairs} Average TF--IDF cosine similarity between explanations produced by models from different AI developers. Values are averaged over all model pairs within each vendor pairing.}
\centering
\begin{tabular}[t]{lrrrr}
\toprule
Vendor Pair & Model Pairs & Mean & Median & SD \\
\midrule
Anthropic–Anthropic & 3 & 0.614 & 0.613 & 0.034 \\
Anthropic–Google & 9 & 0.533 & 0.549 & 0.067 \\
Anthropic–OpenAI & 18 & 0.511 & 0.524 & 0.055 \\
OpenAI–OpenAI & 15 & 0.511 & 0.541 & 0.090 \\
Anthropic–xAI & 9 & 0.493 & 0.485 & 0.081 \\
\addlinespace
Google–Google & 3 & 0.491 & 0.485 & 0.082 \\
Google–OpenAI & 18 & 0.462 & 0.470 & 0.081 \\
Google–xAI & 9 & 0.454 & 0.452 & 0.106 \\
OpenAI–xAI & 18 & 0.450 & 0.433 & 0.082 \\
xAI–xAI & 3 & 0.414 & 0.411 & 0.018 \\
\bottomrule
\end{tabular}
\end{table}

Examination of individual vendor pairings revealed additional structure. Anthropic models exhibited the greatest within-vendor similarity (0.614), substantially exceeding the within-vendor similarity observed for OpenAI (0.511), Google (0.491), and xAI (0.414). Among cross-vendor comparisons, Anthropic–Google (0.533) and Anthropic–OpenAI (0.511) exhibited the highest similarities. Notably, the similarity between Anthropic and OpenAI models was essentially identical to the similarity among OpenAI models themselves (both 0.511), suggesting greater diversity of explanatory style within the OpenAI family than within the Anthropic family.

Pairwise lexical similarity was modestly but consistently higher for models from the same developer than for models from different developers. The mean within-vendor similarity was 0.510 (95\% bootstrap CI: 0.488--0.532), compared with 0.481 (95\% bootstrap CI: 0.460--0.502) for between-vendor pairs. The estimated mean difference was 0.0289 (95\% bootstrap CI: 0.0214--0.0366), indicating a small but consistent vendor-specific effect on explanatory style.

The lexical similarity analyses complement the STM results. Although the models shared a common conceptual organization of statistical reasoning, they differed modestly in the language used to express those ideas. These stylistic differences were most consistent within the Anthropic family, whereas OpenAI, Google, and xAI models exhibited greater within-vendor variation.

\section{Summary and discussion} \label{sec:summary}

The principal finding of this study is that contemporary LLMs differ much more in the reliability with which they apply statistical reasoning than in the statistical concepts they invoke. This distinction was apparent only through the multidimensional evaluation framework employed here. Unlike findings reported for several professional and standardized examinations, none of the models approached perfect performance on the statistics assessments examined. Many widely cited evaluations reporting near-perfect performance of large language models have focused on standardized examinations such as the SAT, GRE, LSAT, USMLE, and Bar Exam \cite{taloni, cma, gupta, chat4bar}. Statistics assessments place additional demands on reasoning that may be less amenable to purely linguistic pattern matching. Unlike many standardized examinations, statistics assessments require reasoning under uncertainty, distinguishing closely related statistical concepts, interpreting hypothetical sampling processes, and connecting numerical results to substantive context. As a result, statistics questions often require a level of conceptual understanding that extends beyond selecting the most plausible answer based on language cues alone. This distinction may help explain why current-generation models, despite strong performance on many standardized examinations, continue to exhibit nontrivial error rates on statistics assessments and frequently produce explanations that appear plausible while containing subtle conceptual inaccuracies. In particular, several models produce responses that are procedurally correct but conceptually incomplete or misleading. More broadly, these findings illustrate that structural topic modeling provides a useful framework for studying statistical reasoning in model-generated explanations, complementing traditional accuracy-based evaluations.

Differences among vendors appear to arise primarily in explanatory style and the reliability with which statistical concepts are applied rather than in the underlying statistical concepts invoked during explanation. These findings suggest that future improvements in statistical reasoning may depend less on expanding statistical knowledge than on improving the reliability with which existing knowledge is applied to novel problems.

While some models achieve relatively high correctness rates, their explanations may still reflect underlying conceptual weaknesses. Lexical similarity findings complement the structural topic modeling results and suggest that evaluating generative AI systems in statistics requires attention not only to whether answers are correct, but also to how those answers are constructed and justified. Although all models relied on a common set of statistical reasoning themes, lexical similarity analysis indicates that developers have adopted distinguishable explanatory styles. These stylistic differences are most consistent within the Anthropic family, whereas OpenAI, Google, and xAI models exhibit greater variation both within and across vendors.

This study has several limitations. The evaluation was restricted to four statistics examinations and a zero-shot prompting protocol, and results may differ under alternative prompting strategies or future model releases. In addition, lexical similarity was quantified using TF–IDF cosine similarity, which captures similarities in word usage rather than deeper semantic equivalence. Finally, because LLMs continue to evolve rapidly, the specific performance rankings reported here should be interpreted as a snapshot of current-generation models rather than permanent differences among systems.

The central implication of this study is that statistical reasoning cannot be adequately characterized by response accuracy alone. Structural topic modeling and lexical similarity analysis provided complementary perspectives on the conceptual organization of statistical reasoning and the language used to express that reasoning. Together, these analyses revealed that contemporary LLMs share a common conceptual organization of introductory statistical reasoning while exhibiting modest but detectable differences in explanatory style. As generative AI becomes increasingly integrated into statistics education, evaluating statistical reasoning will require attention not only to whether models produce correct answers, but also to the conceptual organization and explanatory language that underlie those answers.

\section*{Author Contributions}

Monnie McGee conceived of the idea, performed the analyses for accuracy
and coherency, and wrote the article. Julian Cabrera and Mateo Langston
Smith wrote the code that fed the exams to a high-performance computer
and organized the output into a readable format.

\section*{Acknowledgments}

This work was supported in part by the First Year Research Experience
(FYRE) program at Southern Methodist University. The authors thank the
FYRE program for supporting undergraduate participation in this research.

\bibliographystyle{plainnat}
\bibliography{AAA-stat-wileyNJD}

@article{mcgeesadler2025,
    author = {Monnie McGee and Bivin P. Sadler},
    title = {Generative AI Takes a Statistics Exam: A Comparison of Performance Between ChatGPT3.5, ChatGPT4, and ChatGPT4o-mini},
    journal = {Journal of Data Science},
    volume = {23},
    number = {2},
    year = {2025},
    pages = {269--286},
    doi = {10.6339/25-JDS1174},
    issn = {1680-743X},
    publisher = {School of Statistics, Renmin University of China}
}

@article{schwarz2025,
      author = {Joachim Schwarz},
       title = {The use of generative {AI} in statistical data analysis and its impact on teaching statistics at universities of applied sciences},
        year = "2025",
     journal = {Teaching Statistics},
      volume = {47},
      number = {2},
       pages = {118-128}
}

@article{ruta2025,
author = {Michael R. Ruta and Tony Gaidici and Chase Irwin and Jonathan Lifshitz},
title = {ChatGPT for Univariate Statistics: Validation of {AI}-Assisted Data Analysis in Healthcare Research},
journal = {Journal of Medical Internet Research},
year = {2025},
volume = {27},
pages = {e63550},
url = {https://www.jmir.org/2025/1/e63550},
doi = {10.2196/63550}
}

@misc{rcore,
  author		= "{R Core Team}",
  title			= "R: A language and environment for statistical computing",
  version		= "4.3.2",
  year			= "2023",
  month			= "Oct",
  adsurl		= "{https:www.R-project.org}"
}

@article{taloni,
author = {Andrea Taloni and Massimiliano Borselli and Valentina Scarsi and Costanza Rossi and Vincenzo Scorcia and Giuseppe Giannaccare},
title = {Comparative performance of humans versus {GPT-4.0} and {GPT-3.5} in the self-assessment program of {A}merican {A}cademy of {O}phthalmology},
journal = {Scientific Reports},
volume = {13},
number = {18562},
year = {2023},
doi = {10.1038/s41598-023-45837-2}
}

@article{cma,
author = {Dik Wai Anderson Luk and Whitney Chin Tung Ip and Yat-fung Shea},
title = {Performance of {GPT-4} and {GPT-3.5} in generating accurate and comprehensive diagnoses across medical subspecialties},
journal = {Journal of the Chinese Medical Association},
volume = {87},
number = {3},
pages = {259-260}, 
year = {2024},
doi = {10.1097/JCMA.0000000000001064}
}

@misc{gupta,
title = {Chat {GPT} 4 vs {GPT} 3.5: Comparative Analysis of Open{AI} Tools},
author = {Vibha Gupta},
year = {2023},
howpublished = "Almabetter Website",
  month        = "June",
  note         = {\url{https://www.almabetter.com/bytes/articles/chat-gpt-4-vs-gpt-3-5}}
}

@article{chat4bar,
author = {Debra Cassens Weiss},
title = {Latest version of {ChatGPT} aces bar exam with score nearing 90th percentile},
year = {2023},
month = {March},
journal = {{ABA} Journal},
url = {https://www.abajournal.com/web/article/latest-version-of-chatgpt-aces-the-bar-exam-with-score-in-90th-percentile}
}

@misc{ACTMExam,
  author = {{Arkansas Council of Teachers of Mathematics}},
  title = {{Arkansas Council of Teachers of Mathematics Exam}},
  year = {2011},
  howpublished = {\url{http://example.com}},
  note = {Accessed: February, 2024}
}

@misc{cb2011,
  author       = {{College Board}},
  title        = {AP Statistics Exam},
  year         = {2011},
  publisher    = {College Board},
  address      = {New York, NY},
  note         = {Retrieved from https://apcentral.collegeboard.org/}
}

@article{caos,
author = {Robert delMas and Joan Garfield and Ann Ooms and Beth Chance},
year = {2007},
month = {11},
pages = {28-58},
title = {Assessing students’ conceptual understanding after a first course in statistics},
volume = {6},
journal = {Statistics Education Research Journal},
doi = {10.52041/serj.v6i2.483}
}

@article{ellis23,
author = {Amanda R. Ellis and Emily Slade},
title = {A New Era of Learning: Considerations for ChatGPT as a Tool to Enhance Statistics and Data Science Education},
journal = {Journal of Statistics and Data Science Education},
volume = {31},
number = {2},
pages = {128--133},
year = {2023},
publisher = {Taylor \& Francis},
doi = {10.1080/26939169.2023.2223609},
URL = {https://doi.org/10.1080/26939169.2023.2223609}
}

@misc{KhanAcademy2024,
  author = {{Khan Academy}},
  title = {Four Stars for {KhanMigo}: Common Sense Media Rates {AI} Tools for Learning},
  year = {2024},
  howpublished = {\url{https://blog.khanacademy.org/four-stars-for-khanmigo-common-sense-media-rates-ai-tools-for-learning-kp/}},
  note = {Accessed: 2024-05-06}
}

@misc{exams,
 author = {Lakshmi Varanasi },
howpublished = {\url{https://www.businessinsider.com/list-here-are-the-exams-chatgpt-has-passed-so-far-2023-1}},
year = {2023},
title = {{GPT-4} can ace the bar, but it only has a decent chance of passing the {CFA} exams. Here's a list of difficult exams the {ChatGPT} and {GPT-4} have passed.}
}

@misc{readtext,
title = {readtext: Import and Handling for Plain and Formatted Text Files},
author = {Kenneth Benoit and Adam Obeng}, 
year = {2023},
note = {R package version 0.90},
url = {https://CRAN.R-project.org/package=readtext}}

@article{quanteda,
	author = {Kenneth Benoit and Kohei Watanabe and Haiyan Wang and Paul Nulty and Adam Obeng and Stefan M{\"u}ller and Akitaka Matsuo},
	doi = {10.21105/joss.00774},
	journal = {Journal of Open Source Software},
	number = {30},
	pages = {774},
	title = {quanteda: An {R} package for the quantitative analysis of textual data},
	url = {https://quanteda.io},
	volume = {3},
	year = {2018}}

@article{blei2003,
  title={Latent Dirichlet Allocation},
  author={Blei, David M. and Ng, Andrew Y. and Jordan, Michael I.},
  journal={Journal of Machine Learning Research},
  volume={3},
  pages={993--1022},
  year={2003},
  publisher={MIT Press}
}

@inproceedings{yao2024,
author = {Yao, Shunyu and Yu, Dian and Zhao, Jeffrey and Shafran, Izhak and Griffiths, Thomas L. and Cao, Yuan and Narasimhan, Karthik},
title = {Tree of thoughts: deliberate problem solving with large language models},
year = {2024},
publisher = {Curran Associates Inc.},
address = {Red Hook, NY, USA},
booktitle = {Proceedings of the 37th International Conference on Neural Information Processing Systems},
articleno = {517},
numpages = {14},
location = {New Orleans, LA, USA},
series = {NIPS '23}
}

@article{williams2024,
author = {Andrew Williams},
title = {Comparison of generative {AI} performance on undergraduate and postgraduate written assessments in the biomedical sciences},
journal = {International Journal of Education and Technology in  Higher Education},
year = {2024},
volume = {21},
number = {52},
url = {https://doi.org/10.1186/s41239-024-00485-y}
}

@misc{herklotz2025trustllmstutorstudents,
      title={Can we trust LLMs as a tutor for our students? Evaluating the Quality of LLM-generated Feedback in Statistics Exams}, 
      author={Markus Herklotz and Niklas Ippisch and Anna-Carolina Haensch},
      year={2025},
      eprint={2511.04213},
      archivePrefix={arXiv},
      primaryClass={stat.OT},
      url={https://arxiv.org/abs/2511.04213}, 
}

@article{Nascimento2024LLM4DSEL,
  title={LLM4DS: Evaluating Large Language Models for Data Science Code Generation},
  author={Nathalia Nascimento and Everton Guimar{\~a}es and Sai Sanjna Chintakunta and Santhosh Anitha Boominathan},
  journal={ArXiv},
  year={2024},
  volume={abs/2411.11908},
  url={https://api.semanticscholar.org/CorpusID:274141792}
}

@article{roberts2019,
    title = {{stm}: An {R} Package for Structural Topic Models},
    author = {Margaret E. Roberts and Brandon M. Stewart and Dustin Tingley},
    journal = {Journal of Statistical Software},
    year = {2019},
    volume = {91},
    number = {2},
    pages = {1--40},
    doi = {10.18637/jss.v091.i02},
  }

@article{SALTON1988513,
title = {Term-weighting approaches in automatic text retrieval},
journal = {Information Processing and Management},
volume = {24},
number = {5},
pages = {513-523},
year = {1988},
issn = {0306-4573},
doi = {https://doi.org/10.1016/0306-4573(88)90021-0},
url = {https://www.sciencedirect.com/science/article/pii/0306457388900210},
author = {Gerard Salton and Christopher Buckley}
}

@book{manning2008introduction,
  author    = {Manning, Christopher D. and Raghavan, Prabhakar and Sch{\"u}tze, Hinrich},
  title     = {Introduction to Information Retrieval},
  publisher = {Cambridge University Press},
  year      = {2008},
  address   = {Cambridge},
  isbn      = {978-0521865715}
}

\end{document}